\documentclass[preprint]{article}

\usepackage{neurips_2025}

\usepackage[utf8]{inputenc} 
\usepackage[T1]{fontenc}    
\usepackage{hyperref}       
\usepackage{url}            
\usepackage{booktabs}       
\usepackage{amsfonts}       
\usepackage{amsmath}        
\usepackage{nicefrac}       
\usepackage{microtype}      
\usepackage{xcolor}         
\usepackage{multirow}
\usepackage{textcomp}
\usepackage{utfsym}
\usepackage{svg}

\usepackage{graphicx}
\usepackage{subcaption}

\usepackage{authblk}
\title{Enhancing Spatial Reasoning through Visual and Textual Thinking}

\author[1,3]{Xun Liang}
\author[2]{Xin Guo}
\author[2]{Zhongming Jin}
\author[3]{Weihang Pan}
\author[4]{Penghui Shang}
\author[1]{Deng Cai}
\author[3]{Binbin Lin}
\author[2]{Jieping Ye}
\affil[1]{State Key Lab of CAD\&CG, Zhejiang University}
\affil[2]{Alibaba Cloud Computing}
\affil[3]{School of Software Technology, Zhejiang University}
\affil[4]{Xihu, Hangzhou Zhiyuan Research Institute Co., Ltd}

\begin{document}

\maketitle

\begin{abstract}

    The spatial reasoning task aims to reason about the spatial relationships in 2D and 3D space, which is a fundamental capability for Visual Question Answering (VQA) and robotics. Although vision language models (VLMs) have developed rapidly in recent years, they are still struggling with the spatial reasoning task. In this paper, we introduce a method that can enhance Spatial reasoning through Visual and Textual thinking Simultaneously (SpatialVTS). In the spatial visual thinking phase, our model is trained to generate location-related specific tokens of essential targets automatically. Not only are the objects mentioned in the problem addressed, but also the potential objects related to the reasoning are considered. During the spatial textual thinking phase, Our model conducts long-term thinking based on visual cues and dialogues, gradually inferring the answers to spatial reasoning problems. To effectively support the model's training, we perform manual corrections to the existing spatial reasoning dataset, eliminating numerous incorrect labels resulting from automatic annotation, restructuring the data input format to enhance generalization ability, and developing thinking processes with logical reasoning details. Without introducing additional information (such as masks or depth), our model's overall average level in several spatial understanding tasks has significantly improved compared with other models.

\end{abstract}

\section{Introduction}

    Spatial reasoning aims to understand spatial arrangements in 2D and 3D spaces, which is crucial for accurately interpreting complex visual environments\cite{cheng2024SpatialRGPT}. The significance of enhancing the spatial reasoning of the model lies not only in general visual understanding but also has important practical significance for other downstream tasks such as embodied intelligent decision-making, human-computer interaction, and other active fields\cite{driess2023palme,chen2024spatialvlm}. Visual language models (VLMs) have developed rapidly in recent years and made a significant progress in many basic visual tasks, such as image and video understanding\cite{liu2023llava,liu2023improvedllava,li2024llava,li2024llava-onevision, Qwen-VL, Qwen2-VL, Qwen2.5-VL,chen2024expanding,chen2024far,chen2024internvl,achiam2023gpt,hurst2024gpt,grattafiori2024llama,abdin2024phi}. However, spatial reasoning tasks, such as determining the 3D positions of objects or their spatial relationships, are still complex for most modern VLMs to perform effectively\cite{liu2023visualreasoning}. 

    Previous research on spatial reasoning has typically concentrated on explicit spatial scene memories \cite{gervet2023navigating,gordon2018iqa} or spatial scene graphs \cite{hemachandra2014learning,hildebrandt2020scene,wald2020learning,walter2013learning}. In order to address spatial problems presented in the VQA format, these approaches often need to treat the task explicitly as a path-finding problem within the context of such scene graphs. With the explosive growth of visual language models' capabilities, their spatial reasoning capabilities are expected to be substantially improved. Researchers have first focused on constructing many benchmarks and enhancing the spatial understanding ability of large visual models in a data-driven manner\cite{liu2023visualreasoning,chen2024spatialvlm}. At the same time, some researchers\cite{cheng2024SpatialRGPT} focus on integrating more visual elements (such as depth, region, etc.) to enhance the model's understanding of spatial relationships and scales. 

    \begin{figure}[t]
      \centering
      \includegraphics{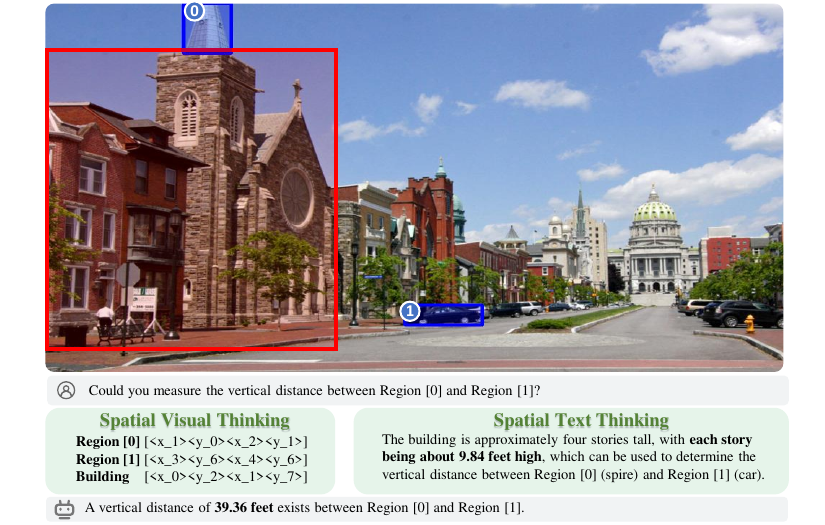}
      \caption{Our model analyzes the original problem and the image to find both the evident and potential visual cues related to the problem in the image and provides corresponding rationales. Based on the visual cues and rationales, our model can effectively enhance the ability of spatial reasoning.}
      \label{fig:demo}
      \vspace{-1.8em}
    \end{figure}

    Our key insight is that spatial reasoning is not merely simple question-and-answer for the targets in the picture. Many spatial reasoning tasks require analysis combined with potential visual cues to give the correct answer. As shown in Fig.\ref{fig:demo}, if we only focus on Region [0] and Region [1], it will be difficult for us to determine their actual vertical distance. In fact, the building where Region [0] is located (indicated by the red box) is an important visual reference in this question. We can first estimate that Region [0] is approximately at the height of a four-story building and then infer the apparent distance between Region [0] and Region [1]. That is to say, visual cues are not limited to the objects mentioned in the question. There are several potential related targets in the image that seem irrelevant to the question but can assist in answering spatial reasoning tasks. In the inference process of VLM, attention should be paid to these potential targets, thereby assisting in the answer to the final question. On the other hand, after obtaining more visual cues, the model should conduct further reasoning on the question and visual cues rather than simply providing a straightforward answer. A large number of visual reasoning works\cite{xu2024llava,liu2025visual,li2025imagine,thawakar2025llamav} have proved that allowing the model to think further can enhance the model's understanding of the task. In the spatial reasoning task, we believe stimulating the model's long-term thinking ability can further explore and utilize visual cues. For the example in Fig.\ref{fig:demo}, textual reasoning can further establish the connections between various visual targets. After further reflection on visual cues, the model finally determine the vertical height between Region [0] and Region [1].

    Inspired by the recent CoT-based\cite{shao2024visual,xu2025llavacotletvisionlanguage} and reasoning-based VLMs\cite{thawakar2025llamav}, we propose a method that can enhance Spatial reasoning through Visual and Textual thinking Simultaneously (SpatialVTS). Our model can be summarized as two phases: Spatial Visual Thinking and Spatial Textual Thinking. In the first phase, the VLM is trained to generate the regional location tokens related to important targets. In the second phase, the VLM takes the question and visual cues into consideration. By stimulating the model's reasoning ability, SpatialVTS can further establish the connections between various visual targets and ultimately determine the final answer. To achieve this goal, we constructe a dataset consisting of questions, answers, relevant target boxes, and rationales. The relevant target boxes are represented using special tokens of discrete positions. The rationales tell us what is in the target box and why it relates to spatial reasoning. Experiments demonstrate that our model can improve spatial reasoning capabilities in average performance across various datasets, even without any assistance of additional visual elements.

    Our contributions can be summarized as follows. 
    
    1. We propose a method that simultaneously enhances spatial reasoning through visual and textual thinking. Spatial visual thinking effectively explores the spatial visual cues and outputs the candidate reference targets. Spatial textual thinking conducts further thinking based on visual cues and provides rationales and the final answer step by step.

    2. To effectively support model training, we analyze the deficiencies of the existing spatial reasoning datasets. We clean and correct the data with relevant target special tokens and rationales through manual annotation.
    
    3. Experimental results demonstrate the effectiveness of our approach. Without incorporating additional information, such as masks or depth data, our model demonstrates a significant improvement in overall average performance across various spatial understanding tasks.

\section{Related Work}


\subsection{Spatial Reasoning Benckmarks}

    Early works on spatial reasoning of large models focus on building the spatial reasoning benchmarks and exploring the performance of the VLMs in spatial reasoning tasks.
    
    Visual Spatial Reasoning (VSR)\cite{liu2023visualreasoning} presents a dataset containing more than 10k natural text-image pairs with 66 types of spatial relations (such as under, in front of, facing). It performs a by-relation analysis and find that the models' performances on certain relations have little correlation with the number of training examples, and certain relations are inherently more challenging. SpatialVLM\cite{chen2024spatialvlm} introduces an automated 3D spatial VQA data generation framework capable of scaling up to 2 billion VQA examples across 10 million real-world images. The authors explore various factors in the training process, including data quality, training pipeline design, and VLM architecture. Training on these datasets can significantly improve models' performance in both qualitative and quantitative aspects of spatial VQA. SpatialRGPT\cite{cheng2024SpatialRGPT} introduces a scalable data pipeline that constructs region-aware spatial reasoning QAs from existing datasets. It presents SpatialRGPT-Bench, a comprehensive benchmark based on ground-truth 3D annotations that span indoor, outdoor, and simulated environments.

\subsection{Visual Elements Enhancement}

    Many researchers consider incorporating more elements into models to enhance the comprehension ability of the vision language model.
    
    Visual CoT\cite{shao2024visual} proposes a novel multi-turn processing pipeline for MLLMs that can dynamically focus on visual inputs and provide intermediate interpretable thoughts. Their work is oriented towards interpretability analysis and does not explicitly enhance spatial vision information. The recent work\cite{yu2025introducing} proposes to use region selection tokens to select the region related to the query and Vision Re-Encoding Tokens to re-encode the original image using re-encode the original image. However, it does not focus on potential reasoning clues and spatial tasks. RegionGPT\cite{guo2024regiongpt} proposes a general framework that harnesses the capabilities of LLMs to tackle complex region-level captioning and understanding tasks. Osprey\cite{yuan2024osprey} proposes a mask-text instruction tuning approach to extend MLLMs by incorporating fine-grained mask regions into language instruction, aiming at achieving pixel-wise visual understanding. For spatial reasoning, SpatialRGPT\cite{cheng2024SpatialRGPT} presents a framework that enhances region-level spatial reasoning in VLMs by enabling effective representation of regional information and acquisition of spatial knowledge. They also integrate depth information flexibly, significantly improving 3D perception and analysis.

\subsection{VLM Reasoning}
    
    The trend of large language models reasoning \cite{jaech2024openai,guo2025deepseek} has quickly spread to the field of visual language models\cite{qvq-72b-preview}. LLaVA-CoT\cite{xu2024llava} independently engages in sequential stages of summarization, visual interpretation, logical reasoning, and conclusion generation. This structured approach enables LLaVA-CoT to achieve marked improvements in precision on reasoning-intensive tasks. Visual-RFT\cite{liu2025visual} introduces a Visual Reinforcement Fine-tuning framework, which adapts the GRPO-based reinforcement learning strategy for enhancing the visual perception and grounding ability of LVLMs. MVoT\cite{li2025imagine} leverages multimodal-native architectures to transcend text-form thinking into multimodal-native reasoning by generating image visualizations of their reasoning traces. This reasoning paradigm enables the model to create reasoning traces and ‘think’ in words and images seamlessly while avoiding the potential errors introduced in captioning the images. LlamaV-o1\cite{thawakar2025llamav} proposes a comprehensive approach for advancing multimodal reasoning by introducing a new benchmark, a novel metric, and an innovative model trained using curriculum learning.

    Our model can be regarded as a region-based and reasoning-based method, which use Spatial Visual Thinking and Spatial Textual Thinking explicitly increase the model's understanding of spatial reasoning.

\section{Spatial Reasoning through Visual and Textual Thinking}

    \begin{figure}[t]
      \centering
      \includegraphics{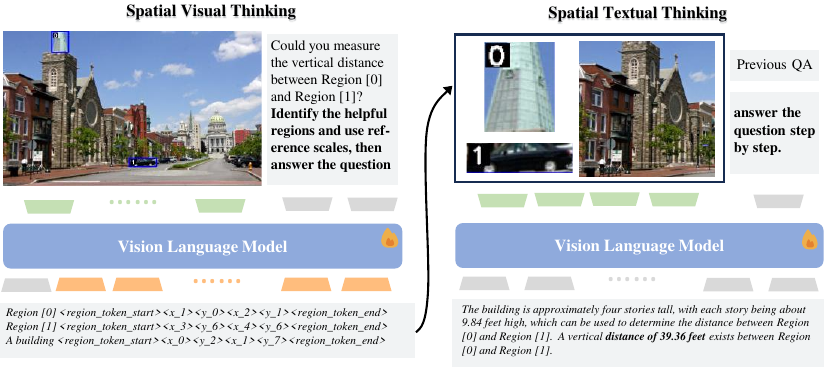}
      \caption{The framework of SpatialVTS. Our model contains two phases: the Spatial Visual Thinking phase generates the region's special token, and the Spatial Textual Thinking phase generates the rationales and the final answer.}
      \label{fig:framework}
    \end{figure}

    Our method can be divided into two stages. The first stage is Spatial Visual Thinking. It analyzes the original problem and the image to find both the evident and potential target positions related to the problem in the image. The second stage is Spatial Textual Thinking. In this stage, the images of the relevant regions are re-input into the VLM as additional visual cues. The model is encouraged to conduct long reasoning based on visual cues and provide the rationales and the final answer.

\subsection{Spatial Visual Thinking}

    Spatial visual thinking aims to identify and extract potential targets related to a given spatial reasoning problem. Unlike most visual VQA questions, where answers can be easily given based on the images, the answers to spatial reasoning tasks (such as the scale of the target, the relative position of the target, etc.) cannot be obtained directly.

\begin{figure}[t]
      \centering
      \includegraphics[width=0.5\linewidth]{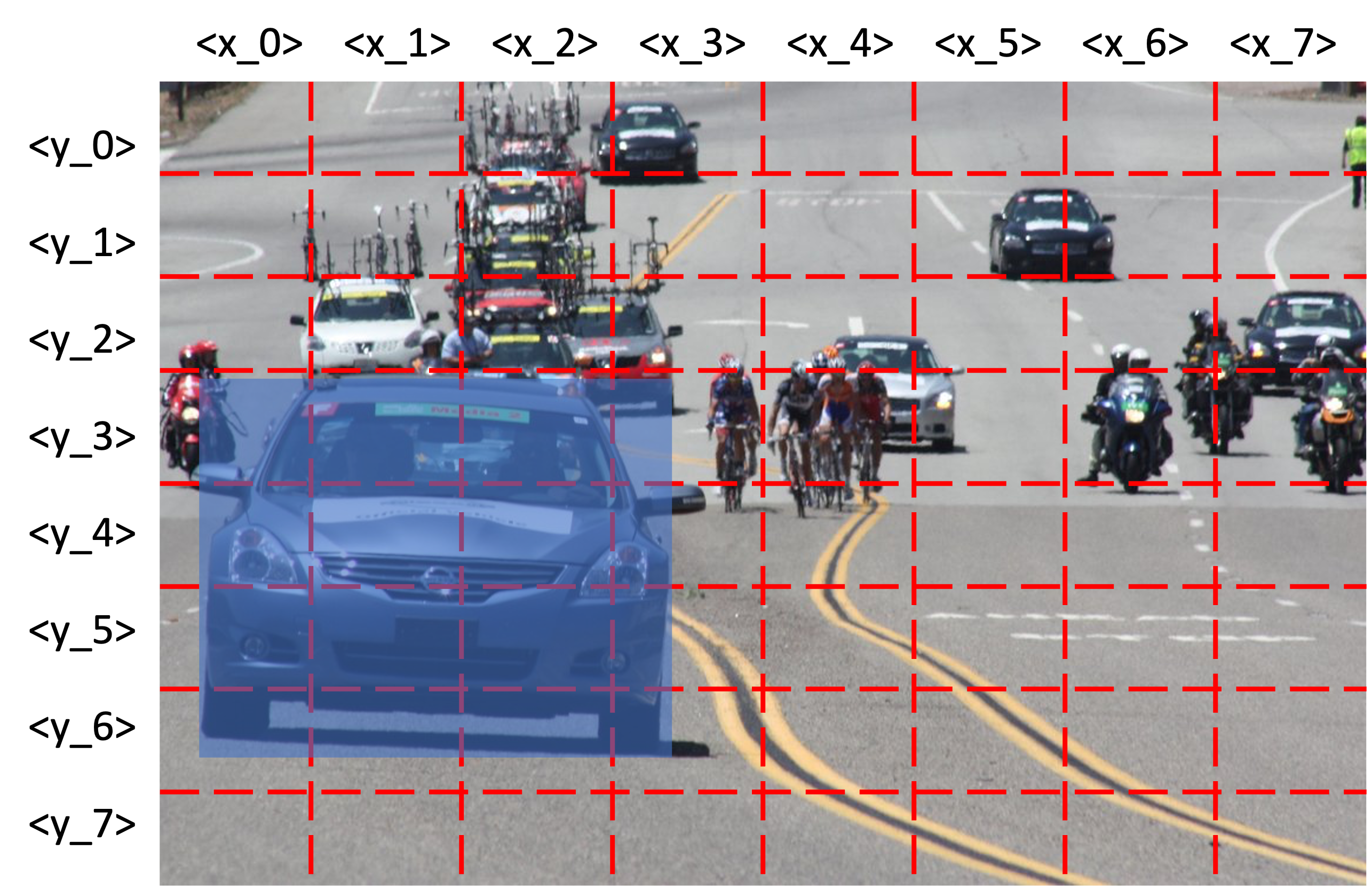}
      \caption{Bounding box representation. We use the indices of the cells containing the top-left and bottom-right pixels of the region $R$ to describe its location. The car in the blue bounding box can be represented as $<x\_0><y\_3><x\_3><y\_6>$.}
      \label{fig:visual_token}
    \end{figure}

    To prevent VLM from struggling to interpret and generate precise bounding boxes, following the work of VPT \cite{yu2025introducing}, we use the region's approximate location to represent the target's position in the image. As in Fig.\ref{fig:visual_token}, we divide the $h \times w$ image evenly into a grid of $k \times k$ rectangular cells, with each cell sized $h/k \times w/k$. Each cell can be indexed by its row and column, with the top-left cell indexed as $(0, 0)$ and the top-right cell as $(k - 1, 0)$. We use the indices of the cells containing the top-left and bottom-right pixels of the region $R$ to describe its location. In our implementation, we set $k = 8$.

    The most significant difference between our method and VPT\cite{yu2025introducing} is that we not only focus on the goals directly mentioned in the problem but also on the potential goals related to spatial reasoning tasks. Secondly, the candidate regions of our method are encoded as visual cues by the same visual encoder and serve as one of the reasoning bases for the next stage. In contrast, the candidate regions in VPT will undergo feature enhancement through other visual feature extractors (such as DINOv2\cite{oquab2023dinov2}) to improve the fine-grained understanding ability of the model. The construction details of the training data will be introduced in detail in Sec.\ref{sec:datasets}.

\subsection{Spatial Textual Thinking}

    For spatial reasoning tasks, logical reasoning is critical. Firstly, the questions of most spatial reasoning tasks are difficult to answer directly. They require analysis in combination with multiple targets in the picture and the overall environment. Secondly, the answers to many spatial reasoning tasks are simple responses such as "up, down, left, right, front or back" and "yes or no". Using overly simplistic answers without the reasoning process in training data will easily cause pattern collapse during the model training process and lead to hallucinations in the final model.
    
    In order to stimulate the thinking ability of the model, we further take the visual cues obtained from the visual encoding of the candidate targets in the previous stage as input. The candidate objects are cropped as new images and inputted into the same vision encoder. Then, we elaborately design a prompt to stimulate the reasoning ability of the model. We use the public visual language model to automatically generate the reasoning process as a part of training data. In particular, we employ the 'seeking the cause by grasping the result' strategy to ensure the rationality and correctness of the generating reasoning process. The question, visual cues, and answer are inputted into the large model simultaneously to find the correct reasoning path. During the training and inference phase, the model is trained to provide rationales and answers only based on the problem and visual cues. 

    \begin{figure}[ht]
        \centering
        \begin{subfigure}[b]{0.48\textwidth}
        \includegraphics[width=\linewidth]{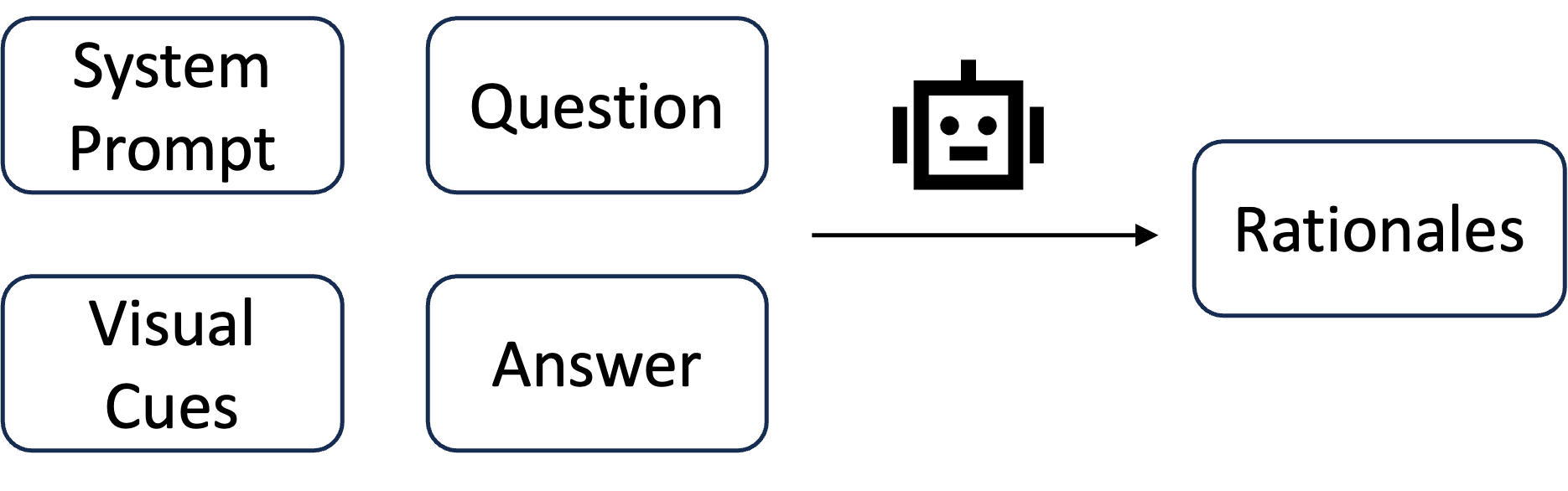}
        \caption{Take the answer as a priori to construct the reasoning process.}
        \label{fig1}
        \end{subfigure}
        \hspace{2mm}
        \begin{subfigure}[b]{0.48\textwidth}
        \includegraphics[width=\linewidth]{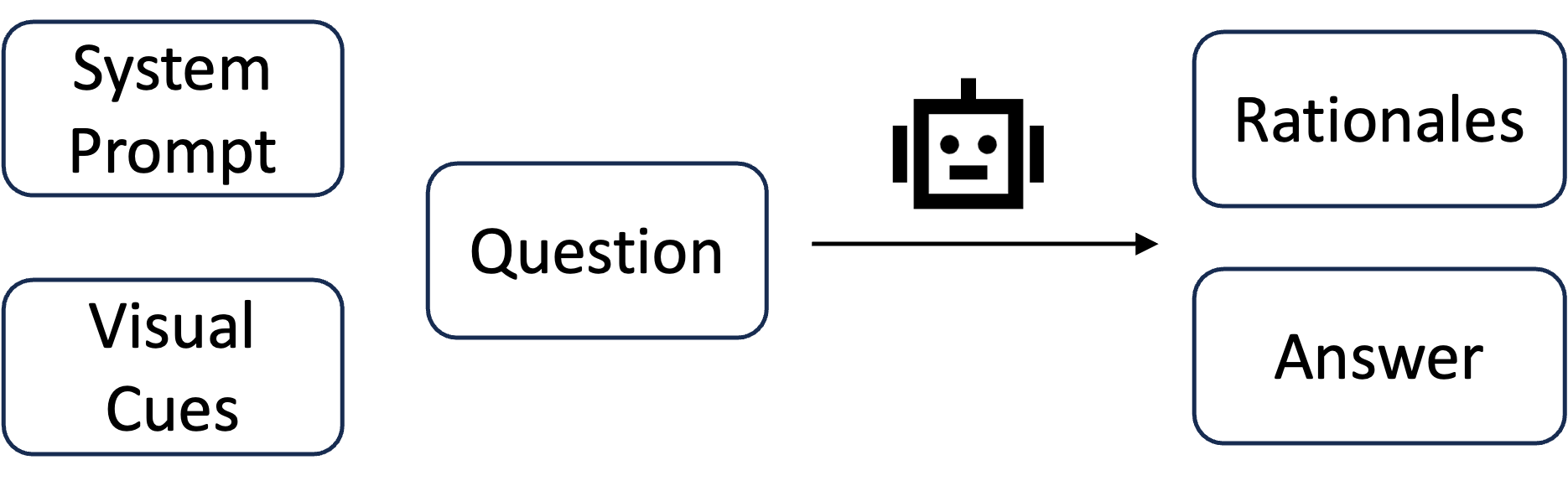}
        \caption{The actual training and inference process of the model.}
        \label{fig2}
        \end{subfigure}
        \caption{Construct the reasoning process via 'seeking the cause by grasping the result'.}
    \label{fig:attention_pooing}
    \end{figure}

\subsection{Dataset Reconstruction}
\label{sec:dataset_reconstruct}

    Our dataset is built based on SpatialRGPT\cite{battaglia2018relational} and VPT\cite{yu2025introducing} dataset. We find that the data needs to be reconstructed to assist the model in deeply exploring the correlations among visual targets.

    1) \emph{The input form is limited.} The SpatialRGPT model is set to require users to explicitly specify masks or bounding boxes as the target's input to eliminate the ambiguity of target reference as much as possible. However, this also limits the model's ability to perform generalized spatial reasoning tasks that only use visual images as input. The SpatialRGPT dataset is tailor-made for the SpatialRGPT model, which means it cannot be directly used for our model training. Therefore, the input form of the data needs to be modified to adapt to the setting of our generalized spatial reasoning problem.
    
\begin{figure}[!t]
      \centering
      \includegraphics[width=\linewidth]{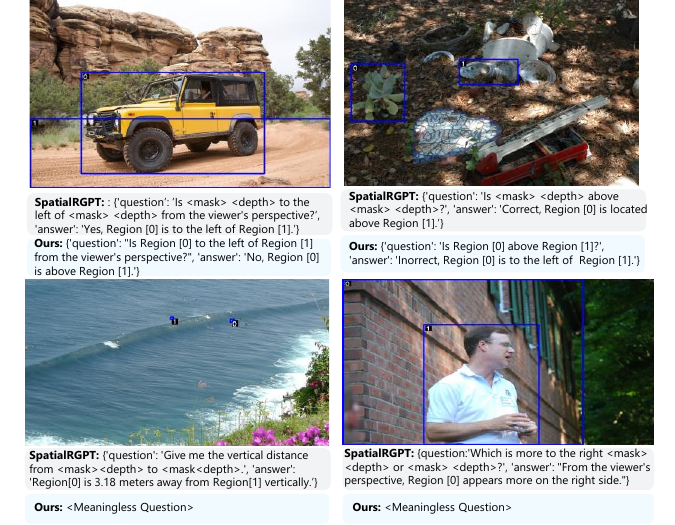}
      \caption{The original Q\&A from SpatialRGPT and our corrected Q\&A.}
      \label{fig:dataset_modify}
    \end{figure}

    2) \emph{The data quality is poor.} The scale of the SpatialRGPT training set is quite large, which is attributed to the pipeline automatically constructed from the data. Nevertheless, the problems based on templating have brought about a series of data quality issues. As in Fig.\ref{fig:dataset_modify}, A considerable proportion of the answers to the questions are incorrect. Meanwhile, since the objects and questions are randomly set according to the template, this has led to a considerable number of questions being uncorrelated with the images. These problems seriously undermine the quality of model training and thus need to be manually checked and corrected.

    3) \emph{Data lacks process-oriented thinking.} The raw SpatialRGPT and VPT dataset provides answers directly based on the questions without intermediate visual or textual thinking. Therefore, based on the original dataset, we cannot obtain the potentially visual targets related to the spatial reasoning task, nor can we know the specific reasoning process.

    Based on the above considerations, for the SpatialRGPT dataset, we replace the <mask> and <depth> placeholder in the user prompt using the corresponding region labels, such as Region [0] and Region [1]. The region bounding boxes are drawn on original images and are marked with the regional tags in the corresponding questions and answers. The above corrections transformed the dataset into the most generalized input format of images and plain text for VLMs. Then, we use manual annotation for data cleaning. First, we filter out the data with unreasonable question settings and correct the incorrect answers. Secondly, we add the bounding box annotations for the potential targets of the problem and provide relevant evidence. Finally, for both the SpatialRGPT and VPT datasets, we use the public vision language model to conduct long-term objective reasoning based on the above-mentioned goals and questions and automatically generate a long-term reasoning process. In particular, we adopt a strategy of 'seeking the cause by grasping the result' to generate the thinking process. We simultaneously input the problems, visual cues, and results into the large model for the reasoning process. This can help the model output the correct reasoning process with high probability.

\section{Experiments}

\subsection{Experiment Setup}
\label{sec:datasets}
    
    \textbf{Training Details} We construct the training dataset to endow VLMs with the ability to answer the spatial reasoning questions quantitatively and qualitatively. We sample 100k samples of data from the original SpatialRGPT dataset\cite{battaglia2018relational}, which contains a vast amount of spatial reasoning data, and 340k samples of data from the VPT dataset\cite{yu2025introducing}, including 140k spatial relation questions and 200k general questions. The merged datasets are then reconstructed following Sec.\ref{sec:dataset_reconstruct}. We choose Qwen2-VL as base model. All training is done using 8 NVIDIA H800 GPUs, with the training time from 7 to 10 hours. We promise to publish our model and datasets after the review is completed.

    \textbf{Benchmarks} To evaluate qualitative performance, we select the most widely adopted benchmarks, including WhatsUP\cite{kamath2023whatsup}, VSR\cite{liu2023visualreasoning}, BLINK-Spatial\cite{fu2024blink}, and the qualitative split of the SRGPT-Bench\cite{battaglia2018relational}, which is denoted as \textit{SRGPT-QUAL}. These benchmarks contain hundreds of spatial relation concepts, such as "up/down" and "front/behind". Meanwhile, we adopt Q-Spatial++\cite{liao2024qspatial} and the qualitative split of the SRGPT-Bench, which is denoted as \textit{SRGPT-QUAN}, to examine the perception of the model of absolute scales. 
    
    
    \textbf{Metrics} To evaluate the model accuracy on qualitative QA benchmarks, we assign a score in $[0,1]$ to answers using DeepSeek-V3-0324\cite{deepseekai2024deepseekv3technicalreport}, and accept answers with scores higher than $0.5$. For quantitative QA benchmarks, the maximum ratio between the ground truth and the answer can be denoted as: $\delta=\max(\frac{\hat{d}}{d^*},\frac{d^*}{\hat{d}})$, where $d^{*}$ is the ground truth and $\hat{d}$ is the estimated distance. Following Q-Spatial++, we accept the answers satisfying $\delta \leq 2$, denoted as $\delta_{\leq2}$. 

    \textbf{Baselines} We compare SpatialVTS with several baselines, including generalized VLMs and VLMs specifically designed for spatial reasoning. \emph{VPT}\cite{yu2025introducing} is a 7B VLM with an additional CLIP encoder trained on OCR, spatial relations, and general QA datasets. \emph{SpaceThinker}\cite{VQASynth}is the newest open-source reproduction of SpatialVLM\cite{chen2024spatialvlm}, and achieves the better performance than SpaceLLaVa\cite{li2024llava}. \emph{SpatialRGPT}\cite{battaglia2018relational} uses more refined depth and mask information to assist in training, while our model is only trained using conventional images and text. Thus, SpatialRGPT is a challenging baseline.

\subsection{Main Results}

    \begin{table}[t]
	\centering
        \setlength\tabcolsep{4pt} 
	\caption{Performance comparison on the qualitative spatial reasoning benchmarks. The numbers in the table represent the accuracy. The best performance is highlighted in bold, and the second performance is underlined. SpatialVTS reaches the best average result.}
    \label{table:main-table}
    \begin{tabular}{lcccccc}
        \toprule
        \multirow{2}{*}{Models} & \multirow{2}{*}{CoT} & \multirow{2}{*}{VSR} & \multirow{2}{*}{SpatialRGPT-QUAL} & \multirow{2}{*}{WhatsUp} & \multirow{2}{*}{BLINK-spatial} & \multirow{2}{*}{Avg.} \\
        & & & & & & \\
        \midrule

        Qwen2-VL-7B\cite{Qwen2-VL} & \usym{2717}  & 63.11 & 51.06 & 82.03 & 57.34 & 63.38 \\
        Qwen2.5-VL-7B\cite{Qwen2.5-VL} & \usym{2717} & 53.96 & 63.38 & \underline{94.66} & 65.03 & 67.26\\
        \midrule
        VPT\cite{yu2025introducing} & \usym{2717} & \textbf{79.00} & 58.45 & 76.45 & \underline{81.18} & 74.47\\
        SpatialRGPT\cite{cheng2024SpatialRGPT} & \usym{2717} & 65.09 & 69.71 & 39.32 & 61.88 & 59.00\\ 
        SpaceThinker\cite{VQASynth} & \usym{2714} & 55.94 & 55.31 & 55.33 & 63.63 & 57.55 \\
        \midrule
        SpatialVTS(vision) & \usym{2717} & 74.50 & \underline{74.87} & 69.17 & \bf{85.31} & \underline{75.96}\\
        SpatialVTS(vision/text) & \usym{2714} & \underline{78.21} & \textbf{78.32} & \textbf{95.38} & 76.22 & \bf{82.03}\\ 
        
    \bottomrule 
    \end{tabular}
    \end{table}

    \begin{figure}[t]
      \centering
      \includesvg[width=\linewidth]{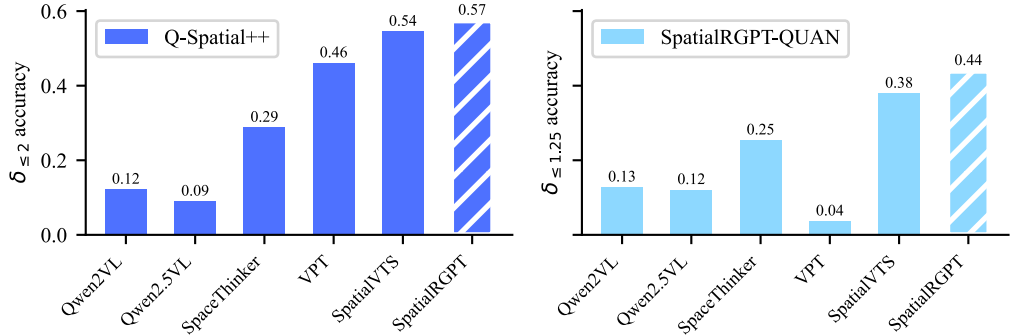}
      \caption{Performance comparison on quantitative spatial QA benchmarks. The white hatch means SpatialRGPT is trained with additional data information (depth and segmentation masks). SpatialVTS outperforms all VLMs trained with text and images, and its performance is almost on par with the SpatialRGPT model that uses depth and mask maps.} 
      \label{fig:qulitative-bench}
    \end{figure}

    Table.\ref{table:main-table} shows the performance comparison on the \textit{qualitative} spatial reasoning benchmarks. We can find that our SpatialVTS achieves the best or suboptimal results in most indicators. In terms of the average effect, our model significantly outperforms other baseline models, proving our method's superiority in spatial reasoning. 
    
    Though some models can reach a high score on a specific benchmark, they can not maintain performance on all tasks. This is because spatial reasoning benchmarks are always vague and biased. A question can have multiple correct answers even from a single perspective, and the observation frame changes a lot. Meanwhile, the answers are always simple, making it easy for VLMs to guess. Therefore, achieving high accuracy on a single benchmark does not mean it substantively understands the spatial relationship. When evaluated on another data distribution, it may make ridiculous mistakes. Only if a model fully understands the spatial concept can it perform well on multiple benchmarks.

    Fig.\ref{fig:qulitative-bench} shows the comparison results on the \textit{quantitative} dataset. On both the Q-Spatial++ and SRPGTBench testing datasets, we find that SpatialVTS outperforms all VLMs trained with text-image pairs, and its performance is nearly on par with SpatialRGPT, which is trained on a large amount of depth and segmentation mask pairs. Since SpatialRGPT utilized depth information during training, it can naturally make more accurate judgments regarding precise distances. However, as the depth information is incorporated into the token list, this comes at the cost of sacrificing part of the model’s QA capabilities, leading to a decline in qualitative metrics. In addition, we notice that all models except VPT demonstrate similar capabilities across the two benchmarks. This is because VPT lacks the concept of units, only providing meaningless numbers on SpatialRGPT-QUAN, and is thus rejected by the evaluator model.

    \begin{table*}[ht]
        \setlength\tabcolsep{4pt}
        \begin{minipage}[b]{0.48\linewidth}
    \small\centering
    \caption{Ablation study of the textual thinking.}
        \label{table:ablation-thinking}
        \begin{tabular}{lcc}
            \toprule  & \text{without text} & \text{with text} \\ 
            \midrule
            \text{SpatialRGPT-QUAN ($\delta_{\leq1.25}$)}& 34.78 & 37.85\\ 
            \text{Q-Spatial++ ($\delta_{\leq2}$)} & 48.10 & 54.45\\
            \bottomrule 
        \end{tabular}
    \end{minipage}
    \hfill
    \begin{minipage}[b]{0.52\linewidth}
    \small\centering
    \caption{Ablation study of the data amount.}
    \label{table:ablation-data}
        \begin{tabular}{l|ccc}
            \toprule  & \text{250k} & \text{440k} \\ 
            \midrule
            \text{VSR    (acc)} & 72.89 & 78.21 \\ 
            \text{Q-Spatial++ ($\delta_{\leq2}$)} & 51.02 & 54.45 \\
            \bottomrule 
        \end{tabular}
    
    \end{minipage}
    \end{table*}

\subsection{Ablation Study}

    
    To verify the effectiveness of the text thinking, we remove the CoT of the training dataset and trained without thinking in text. The results are shown in Table.\ref{table:ablation-thinking}. With text-based reasoning, the performance increases by 3\% and 7\% respectively.
    
    To assess the effectiveness and quality of our training dataset, we construct a tiny dataset with 250k smaples, including 120k CoT samples. Then we compare it to the model trained using full-size dataset. From Table.\ref{table:ablation-data}, even with just the tiny dataset, our model's performance is already comparable to VPT, despite the latter being trained on 800k data points and utilizing an additional visual encoder. Performance can be further improved when more CoT data is used. This validates the effectiveness of our model and dataset.

\subsection{Case Study}
    
\begin{figure}[t]
        \centering
        \includegraphics{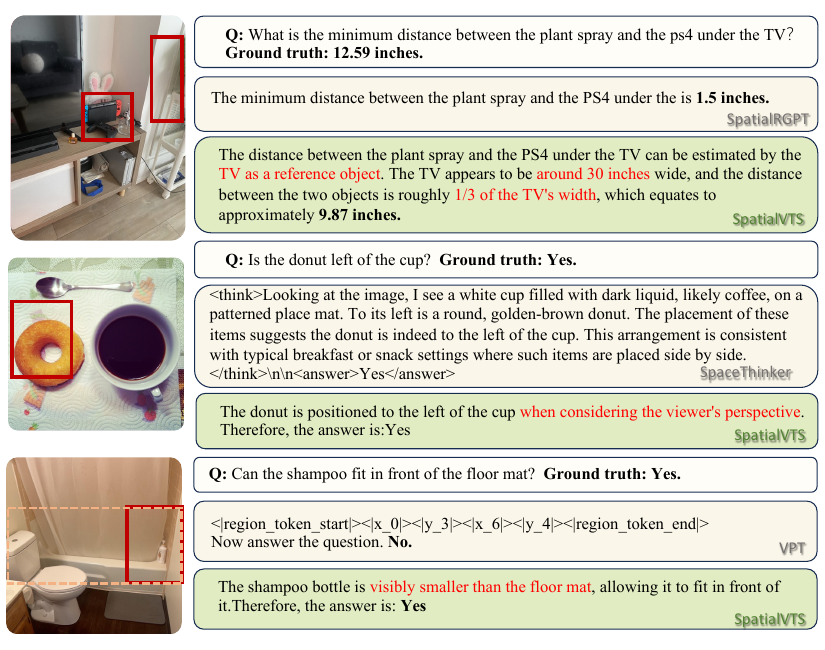}
        \caption{Some reasoning examples. The special regions are bounded with red boxes. SpatialVTS demonstrates the ability to use fine-grained visual grounding and reference scales. The region generated by VPT is bounded by the orange box in the third example.}
        \label{fig:pdf-example}
        \vspace{-1.5em}
    \end{figure}

    
    Fig.\ref{fig:pdf-example} provides a detailed case analysis to give a more concrete perception and understanding of the model effect. In the top case, SpatialRGPT directly provides answers with significant deviations without analysis. At the same time, our model is capable of thoroughly analyzing visual cues and providing logical reasoning and reasonable answers. It uses the common knowledge of the height of the television and uses this as the reference scale to calculate the distance. In the middle case, compared with the SpaceThinker, which provides long but trivial image descriptions, our model can briefly present the key to this problem: the left is relative to the camera's perspective. In the bottom case, both VPT and our model can provide key target boxes, but our model can conduct more in-depth visual cue analysis and give the correct answers. The box given by VPT pays more attention to the 'floor mat' concept, while SpatialVTS can focus on the shampoo and the gap between the mat and the wall.

\section{Conclusion}

    In conclusion, we propose a framework that can enhance spatial reasoning through visual and textual thinking simultaneously. During the spatial visual thinking phase, our model can generate location-related specific tokens of important targets for the question. Then, in the Visual Textual Thinking phase, SpatialVTS takes the question and visual cues into consideration. By stimulating the model's reasoning ability, the model can further establish the connections between various visual targets and ultimately determine the final answer. We filter the quality of the original spatial reasoning data and transform the input format. We further construct a dataset of questions, answers, relevant target boxes, and rationales. Experiments demonstrate that our model can substantially improve the spatial reasoning capabilities with only RGB images as input. In the future, we will further apply our model to embodied intelligent systems to prove the capabilities of the model.

\bibliographystyle{unsrt}
\bibliography{reference}

\end{document}